\documentclass{article}

\usepackage[preprint]{neurips_2026}

\usepackage[utf8]{inputenc} 
\usepackage[T1]{fontenc}    
\usepackage{graphicx}       
\usepackage{wrapfig}        
\usepackage{capt-of}        
\usepackage{hyperref}       
\usepackage{url}            
\usepackage{booktabs}       
\usepackage{amsfonts}       
\usepackage{amssymb}        
\usepackage{nicefrac}       
\usepackage{microtype}      
\usepackage{xcolor}         
\newcommand{\figcell}[1]{\raisebox{-.5\height}{#1}}
\newcommand{\best}[1]{\textbf{#1}}

\newcommand{\Acc}{\mathrm{Acc}}
\newcommand{\PD}{\mathrm{PD}}
\AtBeginDocument{%
  \setlength{\textfloatsep}{8pt plus 2pt minus 2pt}%
  \setlength{\floatsep}{8pt plus 2pt minus 2pt}%
  \setlength{\intextsep}{8pt plus 2pt minus 2pt}%
  \setlength{\abovedisplayskip}{5pt plus 2pt minus 3pt}%
  \setlength{\belowdisplayskip}{5pt plus 2pt minus 3pt}%
  \setlength{\abovedisplayshortskip}{0pt plus 2pt}%
  \setlength{\belowdisplayshortskip}{4pt plus 2pt minus 2pt}%
}

\title{UAV-OVO: Out-of-Viewpoint Generalization in UAV Action Recognition}

\author{%
  Yu Xia \\
  Wuhan University \\
  \And
  Zhengbo Zhang \\
  Singapore University of Technology and Design \\
  \And
  Shuaihu Zhang \\
  Wuhan University \\
  \And
  Zhigang Tu\thanks{Corresponding author.} \\
  Wuhan University \\
}

\begin{document}

\maketitle

\begin{abstract}
  UAV action recognition faces a deployment shift that standard benchmarks often obscure: a model trained on UAV footage captured from low-depression viewpoints may be required to recognize the same action classes from high-depression viewpoints. While the action labels remain unchanged, this shift alters body visibility, motion projection, and scene context, encouraging models to rely on viewpoint-specific shortcuts. We introduce UAV-OVO, an Out-of-Viewpoint generalization benchmark for UAV action recognition. UAV-OVO derives view scores from uncalibrated videos, uses a view-isolation band to assign low-depression videos to the training and in-distribution test splits while reserving high-depression videos for out-of-distribution testing, and constructs ID/OOD test sets matched by class distribution so that performance differences reflect viewpoint shift rather than label imbalance. Across representative video recognizers, UAV-OVO reveals a substantial ID/OOD gap: models that fit the low-depression training distribution well often fail to transfer to held-out high-depression views, exposing viewpoint shortcuts hidden by aggregate accuracy. We further propose LATER, LoRA-Anchored Test-time Re-centering, which first adapts the recognizer with Low-Rank Adaptation (LoRA) and then uses the learned LoRA subspace as a semantic anchor for online feature re-centering. Specifically, LATER projects target-domain displacement onto the orthogonal complement of the LoRA subspace before re-centering features, reducing viewpoint-induced drift while preserving task-relevant semantics. Together, UAV-OVO and LATER provide a controlled testbed and a practical adaptation method for viewpoint-robust UAV video understanding.
\end{abstract}

\section{Introduction}
\label{sec:introduction}

UAV action recognition aims to identify human actions from videos captured by unmanned aerial vehicles. It is an important task for public safety monitoring, search-and-rescue, traffic inspection, and aerial embodied intelligence~\cite{Li_2021_CVPR}. Conventional action recognition has advanced through large-scale video datasets and strong spatiotemporal recognizers~\cite{Kuehne2011HMDB,Soomro2012UCF101,Simonyan2014TwoStream,Tran2015C3D,Kay2017Kinetics,Carreira2017I3D,Feichtenhofer2019SlowFast}. UAV action recognition must also handle footage captured by a moving aerial platform. Camera pitch, yaw, distance to the actor, and flight trajectory can vary across deployment. These viewpoint changes alter body visibility, motion projection, actor scale, and scene context. As a result, the same action may appear familiar from a low-depression, oblique UAV view but visually different from a high-depression, near-overhead view; throughout the paper, these terms refer to smaller and larger downward viewing angles, respectively. UAV action recognition therefore requires models that are robust to systematic viewpoint variation, which makes it a challenging video understanding problem.

\begin{figure}[t!]
  \centering
  \setlength{\tabcolsep}{2pt}
  \begin{tabular}{c c c c}
    & "move a box" & "salute" & "apply cream to face" \\
    \figcell{ID}
    & \figcell{\includegraphics[width=0.285\linewidth]{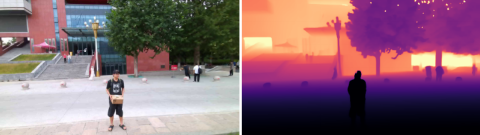}}
    & \figcell{\includegraphics[width=0.285\linewidth]{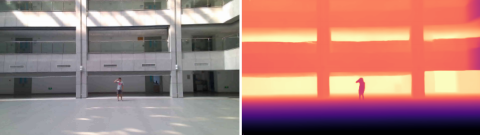}}
    & \figcell{\includegraphics[width=0.285\linewidth]{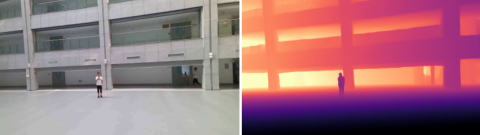}} \\
    \figcell{OOD}
    & \figcell{\includegraphics[width=0.285\linewidth]{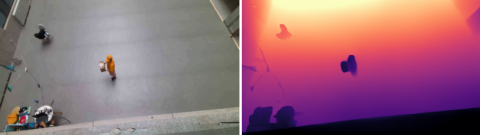}}
    & \figcell{\includegraphics[width=0.285\linewidth]{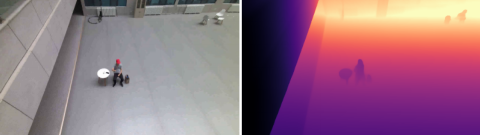}}
    & \figcell{\includegraphics[width=0.285\linewidth]{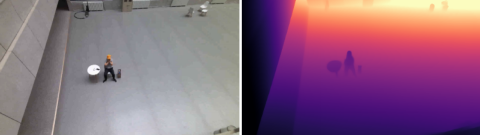}} \\
  \end{tabular}
  \caption{Viewpoint shift in our UAV-OVO benchmark. Columns show the same action class under ID and OOD viewpoints. Each cell contains an RGB frame and the corresponding Depth Anything~v3 (DA3) depth map used to recover scene geometry. Low-depression ID examples preserve more side-view body structure, while high-depression OOD examples compress the actor and motion into a more overhead projection.}
  \label{fig:viewpoint_examples}
\end{figure}

Yet existing evaluation protocols rarely focus on robustness to viewpoint shift. UAV action benchmarks typically report aggregate accuracy on fixed or random splits, where camera viewpoint is not controlled independently from subject identity, scene layout, trajectory, illumination, and class frequency~\cite{Li_2021_CVPR,Kothandaraman2022FAR,Wang2023AZTR}. The visual gap in Figure~\ref{fig:viewpoint_examples} shows why this matters: the same action can preserve side-view body structure in low-depression examples but collapse into a compact overhead projection in high-depression examples. To evaluate this failure mode directly, we introduce UAV-OVO, a UAV Out-of-Viewpoint generalization benchmark. UAV-OVO treats viewpoint shift as an OOD problem with a shared label space: models train on low-depression UAV videos, are evaluated on an in-distribution (ID) low-depression test set, and are separately tested on a held-out high-depression out-of-distribution (OOD) split. This protocol isolates whether recognition performance remains stable when the viewpoint changes while the action classes remain fixed.

Constructing this benchmark requires viewpoint annotations, but existing UAV action datasets do not provide calibrated camera pose for every sample~\cite{Li_2021_CVPR}. UAV-OVO therefore recovers a geometry-derived view score from uncalibrated videos. Specifically, we use DA3 depth and camera-pose predictions~\cite{Lin2025DepthAnything3} to estimate the relation between the camera optical axis and the ground normal, producing a scalar score whose larger values indicate higher depression angles. To reduce frame-level noise, we aggregate scores at the timestamp level, manually review the groups, and use the per-group median score for splitting.

UAV-OVO instantiates a controlled viewpoint-shift evaluation for 155-class UAV action recognition. It uses videos with view scores in \(0\sim30^\circ\) for training and ID testing, removes videos in \(30\sim40^\circ\) as a view-isolation band, and holds out videos above \(40^\circ\) as the OOD test set. To obtain a class-balanced high-depression evaluation set, we additionally collect 981 high-depression videos curated with the same action taxonomy and annotation procedure. This matching controls for label-frequency differences, enabling a more direct measurement of viewpoint-induced degradation rather than class-imbalance effects. Overall, UAV-OVO contains 23,347 videos across 155 action classes, including the samples reserved for the view-isolation band and matched ID/OOD test sets of 3,100 videos each.

Experiments on UAV-OVO show that strong video recognizers still suffer a clear performance drop under viewpoint shift. For example, MViTv2-B~\cite{Li2022MViTv2} reaches 66.45\% accuracy on the ID split but drops to 33.94\% on the high-depression OOD split. Full fine-tuning also remains brittle: it can fit the training set well, yet still fails to preserve accuracy under the held-out viewpoint shift. This pattern suggests training-view over-adaptation rather than simple underfitting. The model does not simply lack UAV-domain adaptation; rather, it tends to entangle action semantics with training-view-specific appearance. A desirable test-time correction should therefore adjust viewpoint-induced feature displacement without overwriting source-adapted action semantics. Motivated by this observation, we introduce LATER, a LoRA-Anchored Test-time Re-centering method. LATER first adapts the recognizer to UAV actions with LoRA and then uses the learned LoRA subspace as a semantic anchor at test time. It corrects source-target feature displacement only in the orthogonal complement of this subspace, preserving source-adapted action semantics while re-centering unlabeled target features without updating the backbone, LoRA weights, or classifier.

This work makes three contributions. First, we formalize out-of-viewpoint generalization for UAV action recognition as a shared-label OOD problem, separating viewpoint robustness from aggregate recognition accuracy. Second, we introduce UAV-OVO, a controlled benchmark built with geometry-derived view scores, timestamp-level viewpoint aggregation, a view-isolation band, and ID/OOD test sets matched by class distribution. Third, we reveal training-view over-adaptation in representative video recognizers and propose LATER, a label-free and gradient-free test-time re-centering method that uses the LoRA subspace as a semantic anchor.

\section{Related work}
\label{sec:related_work}

\textbf{UAV action recognition and distribution-shift benchmarks.} UAV action recognition extends video understanding to moving aerial cameras. Aerial datasets have supported tracking, trajectory forecasting, object detection, overhead imagery, and action understanding from UAV videos~\cite{Robicquet2016StanfordDrone,Mueller2016UAV123,Zhu2018VisDrone,Weir_2019_ICCV,Barekatain2017Okutama,Perera2019DroneAction}. UAV-Human introduced a large-scale benchmark for human behavior understanding from unmanned aerial vehicles and supported several recognition and localization tasks~\cite{Li_2021_CVPR}. Subsequent work improved aerial video representations and temporal reasoning for UAV action recognition~\cite{Kothandaraman2022FAR,Wang2023AZTR}. General video recognizers have also evolved from two-stream and 3D convolutional models toward transformer and masked-pretraining designs~\cite{Wang2016TSN,Tran2018R2Plus1D,Bertasius2021TimeSformer,Arnab2021ViViT,Fan2021MViT,Liu2022VideoSwin,Tong2022VideoMAE}. These studies establish UAV videos as a distinct recognition setting, where actor scale, camera motion, and background context differ from conventional ground-view videos. Existing UAV action protocols, however, mainly report aggregate accuracy on fixed or random splits, leaving it unclear whether models remain reliable under systematic viewpoint changes.

Robustness and distribution-shift benchmarks show that aggregate accuracy can hide large gaps between in-distribution and out-of-distribution performance. Earlier work revealed dataset bias and natural distribution changes in visual recognition~\cite{Torralba2011DatasetBias,Recht2019ImageNetV2,Barbu2019ObjectNet}. Benchmarks such as PACS, DomainNet, and WILDS study domain or environment shifts across visual styles, datasets, and real-world deployment conditions~\cite{Li2017PACS,Peng2019DomainNet,Koh2021WILDS}, while other benchmarks evaluate robustness to corruptions, natural adversarial examples, rendition shifts, and spatial shifts in detection or remote-sensing settings~\cite{Hendrycks2019ImageNetC,Hendrycks2021ImageNetA,Hendrycks2021ManyFacesRobustness,Mao2023COCOO,AlEmadi2025RWDS}. These benchmarks motivate controlled evaluation along a specific deployment-relevant shift variable. UAV-OVO brings this principle to UAV action recognition by defining viewpoint, specifically the change from low-depression to high-depression UAV views, as the controlled shift variable under a shared action label space.

\textbf{Test-time adaptation.} Test-time adaptation uses unlabeled target data to adjust a model during inference. Test-time training introduced the idea of using an auxiliary objective on test samples to improve robustness under distribution shift~\cite{pmlr-v119-sun20b}. Later methods adapted normalization statistics, transferred source hypotheses, or optimized prediction-level objectives such as entropy minimization~\cite{Li2018AdaBN,Liang2020SHOT,Wang2021TENT}. Continual and efficient variants improve stability under non-stationary streams or filter unreliable samples before adaptation~\cite{Wang2022CoTTA,Niu2022EATA,Niu2023SAR}. Other methods use augmentation consistency, classifier adjustment, or latent re-centering to reduce test-time shift without source data~\cite{Zhang2022MEMO,Iwasawa2021T3A,Murphy2025NEO}. These approaches can be effective, but updating model parameters during online video evaluation may be costly or unstable. UAV target videos arrive sequentially, their viewpoints can change with flight trajectory, and their labels are unavailable. LATER keeps all model parameters fixed at test time. It uses target videos only to estimate an online feature displacement, then corrects target features through gradient-free re-centering.

\textbf{Parameter-efficient adaptation.} Parameter-efficient adaptation specializes a pretrained model without updating all weights. Adapter modules, bias-only updates, prompt tuning, and vision-specific adapters reduce the number of trainable parameters while preserving most pretrained weights~\cite{Houlsby2019Adapters,BenZaken2022BitFit,Jia2022VPT,Chen2022AdaptFormer}. LoRA represents the update to a weight matrix as a low-rank residual~\cite{Hu2022LoRA}. This residual is usually treated as an efficient training mechanism, but its low-rank structure also has a geometric interpretation: after source adaptation, it identifies a restricted set of feature directions that the model used to encode task-specific semantic changes. LATER uses this update subspace as a semantic anchor rather than only a parameter-saving device. At test time, viewpoint correction is applied in the orthogonal complement of the LoRA subspace, so feature re-centering can reduce target-view displacement while avoiding directions that source adaptation used for action discrimination. This links parameter-efficient source adaptation with label-free viewpoint correction.

\begin{table}[htbp]
  \caption{Comparison with related aerial action and distribution-shift benchmarks. UAV-OVO combines UAV video action recognition with a controlled viewpoint-shift OOD protocol.}
  \label{tab:benchmark_comparison}
  \centering
  \small
  \setlength{\tabcolsep}{4pt}
  \begin{tabular}{@{}p{0.38\linewidth}cccc@{}}
    \toprule
    Benchmark & UAV video & Action task & OOD protocol & View OOD \\
    \midrule
    Okutama-Action~\cite{Barekatain2017Okutama} & \checkmark & \checkmark & \(\times\) & \(\times\) \\
    Drone-Action~\cite{Perera2019DroneAction} & \checkmark & \checkmark & \(\times\) & \(\times\) \\
    UAV-Human~\cite{Li_2021_CVPR} & \checkmark & \checkmark & \(\times\) & \(\times\) \\
    Image OOD benchmarks~\cite{Li2017PACS,Peng2019DomainNet,Koh2021WILDS} & \(\times\) & \(\times\) & \checkmark & \(\times\) \\
    Detection OOD benchmarks~\cite{Mao2023COCOO,AlEmadi2025RWDS} & \(\times\) & \(\times\) & \checkmark & \(\times\) \\
    UAV-OVO & \checkmark & \checkmark & \checkmark & \checkmark \\
    \bottomrule
  \end{tabular}
\end{table}

\begin{figure}[t!]
  \centering
  \includegraphics[width=0.96\linewidth]{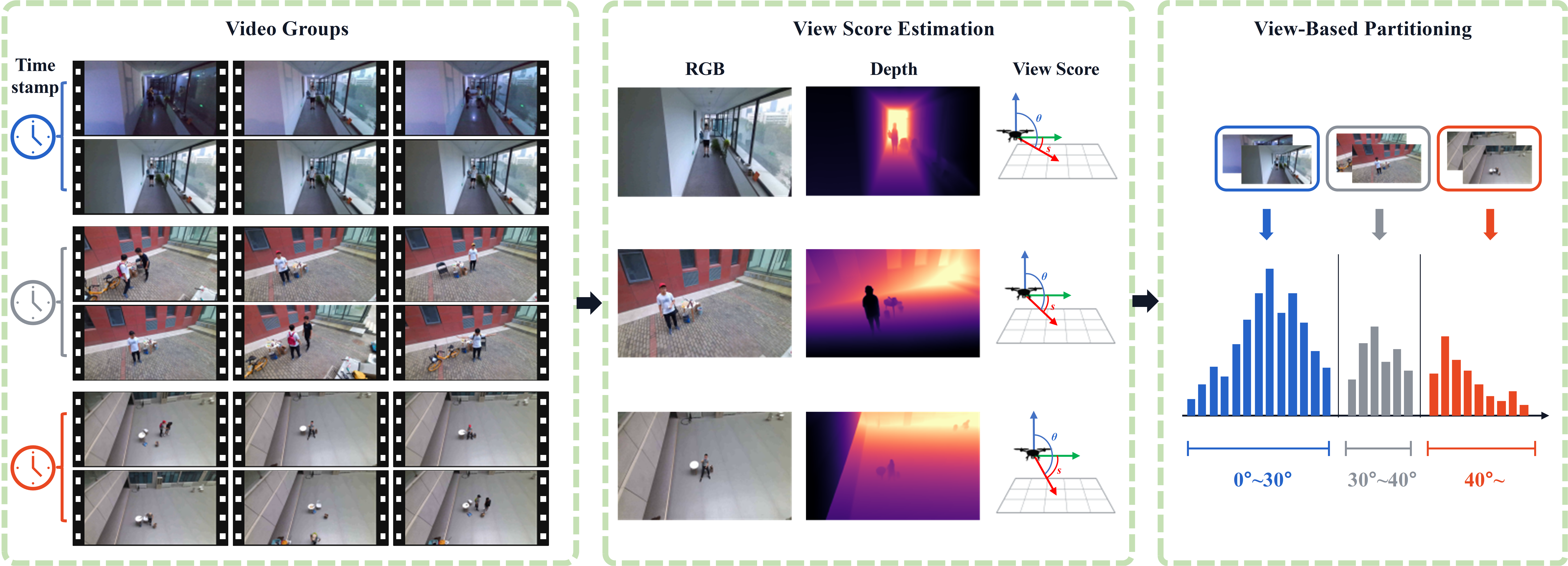}
  \caption{UAV-OVO construction. Timestamp groups receive DA3-derived view scores \(s=\theta-90^\circ\) and are split into low-depression training/ID, isolation, and high-depression OOD regimes under a shared action label space.}
  \label{fig:uav_ovo_construction}
\end{figure}

\section{Method}
\label{sec:method}
\paragraph{Problem setup.}
Given labeled source videos \(\mathcal{D}_s=\{(x_i^s,y_i^s)\}\) from low-depression UAV views, we evaluate a recognizer on two test sets with the same action label space: a low-depression in-distribution set \(\mathcal{D}_{\mathrm{ID}}\) and a held-out high-depression out-of-distribution set \(\mathcal{D}_{\mathrm{OOD}}\). UAV-OVO defines the viewpoint-based split, while LATER adapts the source-trained recognizer to unlabeled target videos from \(\mathcal{D}_{\mathrm{OOD}}\) without updating model parameters at test time.
\subsection{UAV-OVO benchmark}
\label{sec:uav_ovo}

UAV-OVO turns viewpoint from an uncontrolled nuisance variable into the axis of an OOD evaluation. We build it on UAV-Human~\cite{Li_2021_CVPR}, which provides large-scale UAV videos but does not annotate calibrated camera pose for each sample. We use a geometric view score for aerial viewpoint. For each valid frame, we compute the raw angle \(\theta\) between the camera optical axis and the estimated ground normal, then subtract \(90^\circ\) to obtain a pitch offset \(s=\theta-90^\circ\). This score captures how far the camera deviates from a level-view reference toward a high-depression viewpoint and matches the role of camera pitch in monocular 3D human analysis~\cite{Zhan_2022_CVPR}.

We estimate this score directly from video content with a pose-capable DA3 model~\cite{Lin2025DepthAnything3}. For each video, we sample frames at 4 FPS and obtain dense depth, camera intrinsics, and world-to-camera extrinsics. We invert the extrinsics to camera-to-world poses \(T_{cw}=(R_{cw},t_{cw})\), define the optical axis as \(\mathbf{o}=R_{cw}[0,0,1]^\top\), and normalize \(\mathbf{o}\). We back-project DA3 depth to 3D with the predicted intrinsics on a stride-8 pixel grid. Candidate ground points are drawn from the lower central image region, with very near or far depths removed and high depth-gradient pixels filtered out. We fit planes to these candidates with 200-iteration RANSAC; frames with fewer than 32 candidate points or fewer than 16 final plane inliers are invalid. The plane normal is oriented in camera space so its \(y\)-component is non-positive and is then transformed to world coordinates as \(\mathbf{n}\). A valid frame therefore has finite DA3 geometry and an accepted ground plane. We compute \(\theta=\arccos(\mathrm{clip}(\mathbf{o}^{\top}\mathbf{n},-1,1))\) in degrees and use \(s=\theta-90^\circ\) as the frame score. The video score is the median over valid frames. Single-video estimates can still be noisy because the actor, shadows, occlusions, or non-planar background may affect depth and plane fitting. UAV-Human filenames encode year, month, hour, and minute, and videos recorded at the same timestamp usually share nearly identical camera viewpoints. We therefore group videos by timestamp, manually review the groups, and assign each group the median video score. Manual review only verifies timestamp groups and removes obvious geometry-estimation failures; it does not use action labels, model predictions, or ID/OOD performance. This aggregation reduces per-video estimation noise while preserving the viewpoint structure needed for splitting.

The split separates viewpoint regimes rather than random samples. Videos with view scores in \(0\sim30^\circ\) form the low-depression pool used for training and ID testing. Videos in \(30\sim40^\circ\) are removed as an isolation band, which reduces boundary ambiguity and prevents near-threshold samples from weakening the shift. Videos above \(40^\circ\) form the held-out high-depression OOD test set. Figure~\ref{fig:uav_ovo_construction} shows the complete construction pipeline, from timestamp grouping to viewpoint-based splitting. Since class imbalance can mimic or obscure robustness failures, we further collect 981 high-depression videos to match the ID test distribution. This top-up contains 967 videos recorded with a DJI Flip under the same UAV action-recognition setup and 14 manually filtered web videos. We use the added videos only for OOD testing, not for training. After this top-up, each of the 155 action classes has the same number of ID and OOD test videos. The final split statistics are reported in Table~\ref{tab:uav_ovo_splits}. The isolation band is part of UAV-OVO, but it is excluded from the main ID/OOD accuracy calculation. UAV-OVO evaluates each method on ID accuracy and OOD accuracy under the same label space. This protocol reports ID and OOD accuracy separately, making viewpoint-induced degradation visible rather than folding it into a single aggregate score.

\begin{table}[t!]
  \caption{UAV-OVO split statistics. The ID and OOD test sets contain the same 155 action classes and the same number of videos, while differing in view-score range.}
  \label{tab:uav_ovo_splits}
  \centering
  \small
  \begin{tabular}{lccc}
    \toprule
    Split & View-score range & Videos & Classes \\
    \midrule
    Train & \(0\sim30^\circ\) & 13,872 & 155 \\
    ID test & \(0\sim30^\circ\) & 3,100 & 155 \\
    Isolation band & \(30\sim40^\circ\) & 3,275 & 155 \\
    OOD test & \(>40^\circ\) & 3,100 & 155 \\
    \midrule
    Total & -- & 23,347 & 155 \\
    \bottomrule
  \end{tabular}
\end{table}

\begin{figure}[t!]
  \centering
  \includegraphics[width=0.94\linewidth]{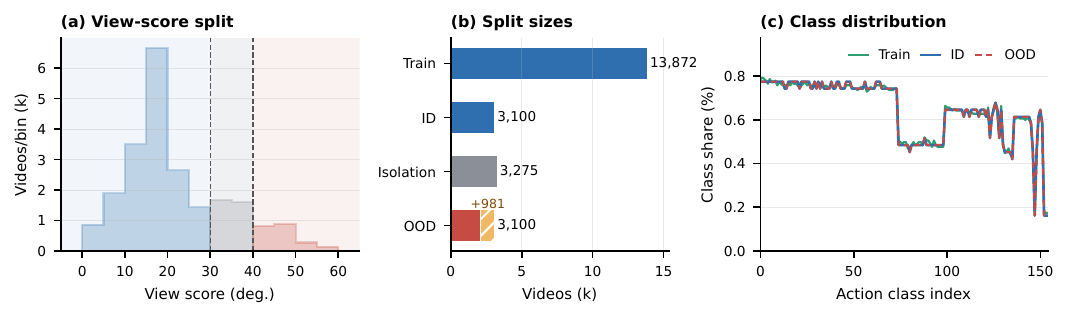}
  \caption{UAV-OVO split distribution. View scores separate low-depression training/ID, isolation, and high-depression OOD regimes; split-size bars highlight the 981-video top-up; class profiles show the matched ID/OOD label distribution.}
  \label{fig:uav_ovo_distribution}
\end{figure}

The split distribution provides three checks on the benchmark construction, as shown in Figure~\ref{fig:uav_ovo_distribution}. After timestamp-level review, the view-score panel places each video in its final split regime, making the source, isolation, and target intervals visually disjoint. The split-size bars make the final OOD composition explicit, including the 981 high-depression top-up videos. The normalized class profiles show that Train, ID, and OOD follow nearly the same per-class distribution, with an exact ID/OOD test match. This matching reduces label imbalance as a competing explanation for the measured OOD gap.

\begin{figure}[t!]
  \centering
  \scriptsize
  \setlength{\tabcolsep}{1.4pt}
  \begin{tabular}{@{}c c c c c c@{}}
    & "salute" & \shortstack{"apply cream\\to face"} & \shortstack{"apply cream\\to hands"} & "open a box" & "move a box" \\
    \figcell{\shortstack{UAV-Human\\OOD}}
    & \figcell{\includegraphics[width=0.160\linewidth]{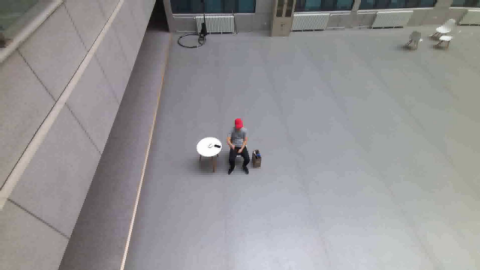}}
    & \figcell{\includegraphics[width=0.160\linewidth]{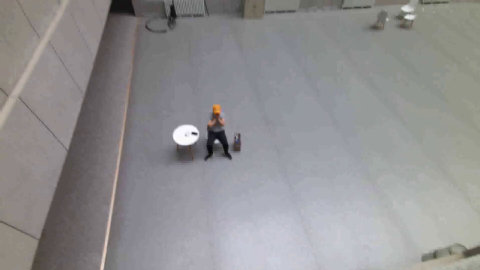}}
    & \figcell{\includegraphics[width=0.160\linewidth]{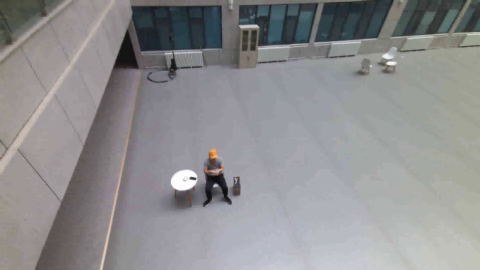}}
    & \figcell{\includegraphics[width=0.160\linewidth]{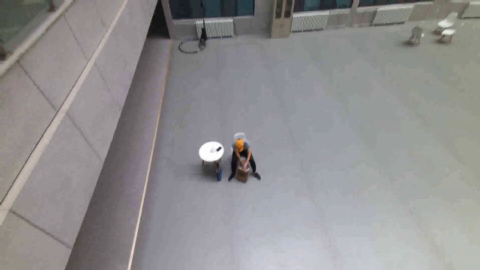}}
    & \figcell{\includegraphics[width=0.160\linewidth]{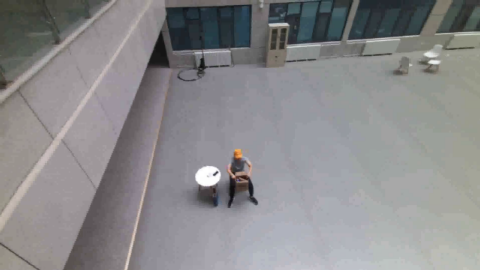}} \\
    \figcell{\shortstack{Top-up\\OOD}}
    & \figcell{\includegraphics[width=0.160\linewidth]{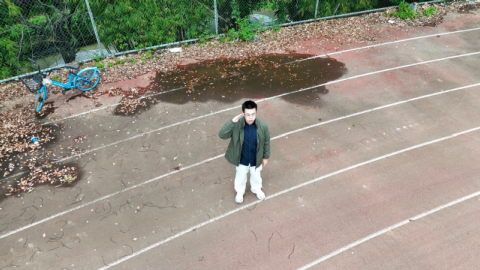}}
    & \figcell{\includegraphics[width=0.160\linewidth]{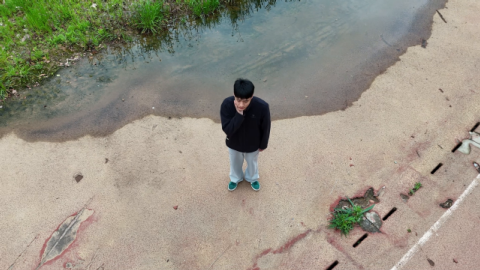}}
    & \figcell{\includegraphics[width=0.160\linewidth]{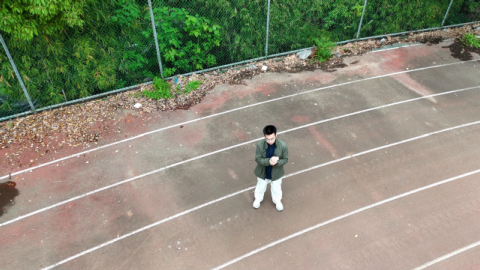}}
    & \figcell{\includegraphics[width=0.160\linewidth]{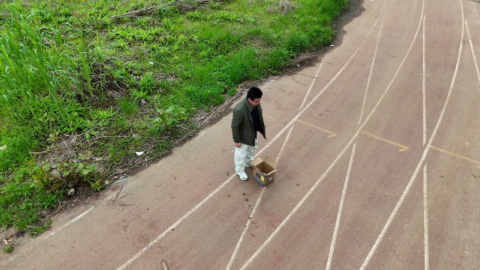}}
    & \figcell{\includegraphics[width=0.160\linewidth]{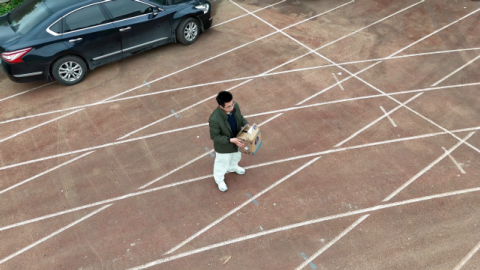}} \\
  \end{tabular}
  \caption{OOD top-up visual check. Each column shows one action class from the original UAV-Human OOD set and the corresponding top-up set. DA3 estimates the top-up examples at \(44.8^\circ\), \(59.0^\circ\), \(45.3^\circ\), \(45.4^\circ\), and \(46.5^\circ\), satisfying the high-depression criterion.}
  \label{fig:topup_ood_examples}
\end{figure}

The top-up examples in Figure~\ref{fig:topup_ood_examples} visually check whether the added videos match the high-depression OOD criterion. The added videos are not identical to UAV-Human in background or hardware, but they preserve the UAV overhead domain, the action definitions, and the high-depression viewpoint required by the OOD split.

\subsection{LATER: LoRA-anchored test-time re-centering}
\label{sec:later}

\begin{figure}[htbp]
  \centering
  \includegraphics[width=0.98\linewidth]{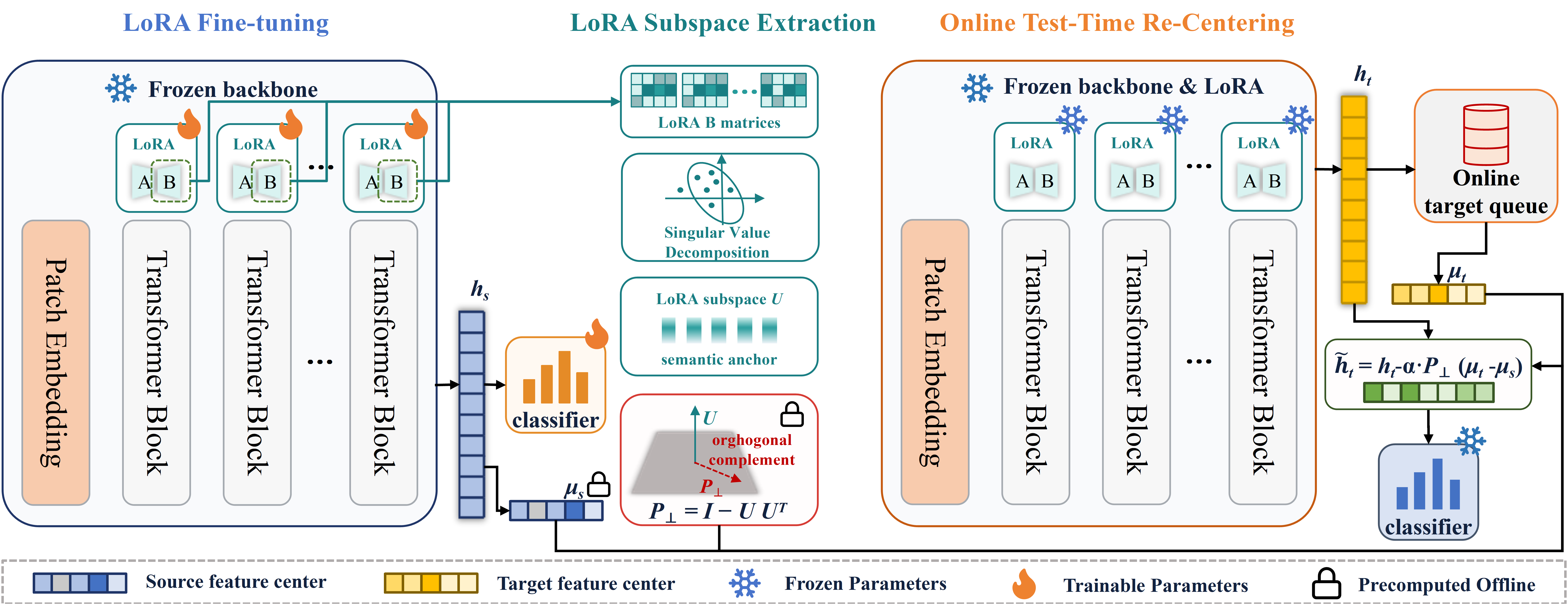}
  \caption{LATER pipeline. Source training learns LoRA adapters and a classifier with the backbone frozen. Offline LoRA-\(B\) matrices define a feature-space semantic anchor. Online LATER re-centers features in the LoRA-orthogonal complement without test-time parameter updates.}
  \label{fig:later_overview}
\end{figure}

LATER addresses the feature displacement induced by viewpoint shift. Let \(f_\theta\) be a video encoder and \(g_\phi\) be a classifier. For a video \(x\), the model extracts a pooled feature \(h=f_\theta(x)\) and predicts with \(g_\phi(h)\). Latent re-centering methods show that distribution shifts can move target embeddings away from the source feature center, and that subtracting an estimated centroid shift can improve alignment with a source classifier~\cite{Murphy2025NEO}. UAV viewpoint shift has a similar feature-space structure, but the correction must preserve action-discriminative directions learned during source adaptation. LATER therefore uses LoRA as both a source adaptation mechanism and a geometric anchor for test-time re-centering; Figure~\ref{fig:later_overview} illustrates this source-adaptation and target-correction pipeline.

During source training, LATER adapts the video recognizer with LoRA~\cite{Hu2022LoRA}. For a pretrained weight matrix \(W\), LoRA represents the adapted weight as
\[
    W' = W + \frac{\gamma}{r}BA,
\]
where \(A\) and \(B\) are low-rank matrices with rank \(r\), and \(\gamma\) controls the update scale. We update the LoRA parameters and the classifier on labeled source videos while keeping the pretrained backbone weights fixed. This restricted update lets the model adapt to UAV action labels without giving the full backbone enough freedom to reshape all feature directions around training-view cues.

After source adaptation, the learned LoRA matrices define source-adaptation directions in feature coordinates. Let the pooled feature dimension be \(d\), so that \(h,\mu_s,\mu_t\in\mathbb{R}^{d}\) and the classifier \(g_\phi\) acts on this final feature space. For each LoRA adapter \(\ell\), we use its \(B_\ell\) matrix only when its output dimension matches \(d\), giving \(B_\ell\in\mathbb{R}^{d\times r}\). Adapters with other output dimensions are excluded from the feature-space projector. We stack these matrices as
\[
    M = [B_1^\top;\ldots;B_L^\top]\in\mathbb{R}^{Lr\times d}.
\]
Singular value decomposition of \(M\) gives right singular vectors in the same \(d\)-dimensional space as the pooled feature. We retain the \(k\) directions whose singular values exceed \(10^{-4}\) of the largest singular value and write them as \(U=[v_1,\ldots,v_k]\in\mathbb{R}^{d\times k}\). LATER then constructs the orthogonal projector
\[
    P_{\perp} = I_d - UU^\top .
\]
The LoRA subspace approximates the directions used by source-domain adaptation. LATER treats these directions as a semantic anchor: test-time correction should avoid moving features along directions that the source-adapted recognizer uses for action discrimination.

LATER estimates viewpoint displacement online. We first compute a source feature center from training videos,
\[
    \mu_s = \frac{1}{N_s}\sum_{i=1}^{N_s} f_\theta(x_i^s).
\]
During evaluation, target videos arrive sequentially. LATER maintains an online queue \(Q\) of target features. For the current target video, it extracts a designated queue feature, appends it to \(Q\), and estimates the target center as
\[
    \mu_t = \frac{1}{|Q|}\sum_{h\in Q} h .
\]
This update uses only unlabeled target features. In our multi-view implementation, the designated queue feature is the first multi-view feature, while the same correction is applied to all view features before logit averaging. Given the source-target displacement \(\mu_t-\mu_s\), LATER applies the LoRA-anchored correction
\[
    \Delta = \alpha P_{\perp}(\mu_t-\mu_s),
    \qquad
    \tilde{h}=h-\Delta ,
\]
where \(\alpha\) controls the re-centering strength. For multi-clip and multi-crop inference, LATER applies the same \(\Delta\) to each view feature of the target video and averages the resulting logits \(g_\phi(\tilde{h})\).

LATER is label-free and gradient-free at test time. It uses target videos only to estimate \(\mu_t\), and it never updates the backbone, LoRA weights, or classifier during evaluation. Compared with parameter-updating test-time adaptation, this design avoids overwriting source-trained action semantics. If \(P_\perp\) is removed, direct re-centering can shift features along LoRA-adapted semantic directions and weaken source-trained class separation. LATER avoids this by reserving the LoRA subspace and correcting only the residual directions.

\section{Experiments}
\label{sec:experiments}

\subsection{Experimental setup}
\label{sec:experimental_setup}

\begin{table}[t!]
  \caption{Main results on UAV-OVO. The backbone and pretrain columns identify the model used for adaptation or evaluation. \(\PD\downarrow\) and \(H\uparrow\) measure OOD drop and balance.}
  \label{tab:main_results}
  \centering
  \footnotesize
  \setlength{\tabcolsep}{3.2pt}
  \begin{tabular}{@{}llllcccc@{}}
    \toprule
    Method & Venue & Backbone & Pretrain & \(\Acc_{ID}\) & \(\Acc_{OOD}\) & \(\PD\downarrow\) & \(H\uparrow\) \\
    \midrule
    ViT-S~\cite{Dosovitskiy2021ViT} & ICLR'21 & ViT & K400 & 47.71 & 15.58 & 0.67 & 23.49 \\
    X3D-M~\cite{Feichtenhofer2020X3D} & CVPR'20 & X3D & K400 & \textbf{68.45} & 25.45 & 0.63 & 37.10 \\
    FAR~\cite{Kothandaraman2022FAR} & ECCV'22 & X3D & K400 & 54.55 & 13.42 & 0.75 & 21.54 \\
    ViViM~\cite{Chen2026VideoMambaSuite} & IJCV'26 & ViViM & K400 & 66.60 & 28.56 & 0.57 & 39.98 \\
    MViTv2-B~\cite{Li2022MViTv2} & CVPR'22 & MViTv2 & K400 & 66.45 & 33.94 & 0.49 & 44.93 \\
    DejaVid~\cite{Ho2025DejaVid} & CVPR'25 & MViTv2 & K400 & 64.19 & 30.97 & 0.52 & 41.78 \\
    NEO~\cite{Murphy2025NEO} & ICLR'26 & MViTv2 & K400 & 66.71 & 34.58 & 0.48 & 45.55 \\
    \midrule
    LATER & Ours & MViTv2 & K400 & 67.06 & \textbf{35.55} & \textbf{0.47} & \textbf{46.47} \\
    \bottomrule
  \end{tabular}
\end{table}

We evaluate on UAV-OVO using low-depression videos as the source training distribution. The ID and OOD test sets contain the same 155 action classes and the same number of videos, so the evaluation measures viewpoint robustness without confounding the result with label-space or class-frequency mismatch. We compare LATER with general-purpose video recognizers, UAV-specific recognition methods, and recent video baselines. All reported evaluations use a unified multi-view protocol with \(5\) temporal clips and \(3\) spatial crops per video, and we average logits over the resulting 15 views. Unless otherwise stated, LATER uses LoRA rank \(r=16\) and re-centering strength \(\alpha=1.0\), which gives the ID-preserving setting used in the main comparison. Full implementation details are provided in the appendix.

Following distribution-shift benchmarks~\cite{AlEmadi2025RWDS}, we report both ID and OOD performance together with two robustness metrics:
\[
    \PD = \frac{\Acc_{ID} - \Acc_{OOD}}{\Acc_{ID}},
    \qquad
    H = \frac{2\Acc_{ID}\Acc_{OOD}}{\Acc_{ID}+\Acc_{OOD}} .
\]
Here, Performance Drop measures the relative degradation from the ID test set to the OOD test set, while \(H\) summarizes the balance between ID recognition and OOD generalization. Lower \(\PD\) and higher \(H\) indicate better robustness. We report \(\PD\) as a ratio and all accuracies and \(H\) as percentages.

\subsection{Main results}
\label{sec:main_results}

\begin{wrapfigure}{r}{0.50\linewidth}
  \vspace{-6pt}
  \centering
  \includegraphics[width=\linewidth]{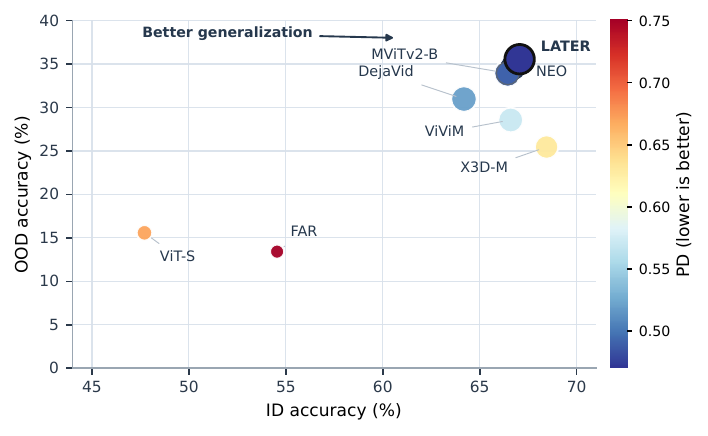}
  \caption{ID/OOD robustness scatter. Marker size encodes \(H\), and color encodes \(\PD\). LATER gives the strongest OOD accuracy and \(H\)-score among the compared methods.}
  \label{fig:id_ood_scatter}
  \vspace{-6pt}
\end{wrapfigure}

The main comparison exposes a substantial viewpoint gap across representative recognizers, as reported in Table~\ref{tab:main_results}. High ID accuracy does not imply viewpoint robustness: X3D-M~\cite{Feichtenhofer2020X3D} obtains the strongest ID accuracy among the listed baselines, but its OOD accuracy drops to 25.45\%. MViTv2-B~\cite{Li2022MViTv2} provides the strongest complete baseline with 66.45\% ID accuracy, 33.94\% OOD accuracy, and an \(H\)-score of 44.93. LATER improves OOD accuracy to 35.55\% and raises \(H\) to 46.47 while preserving competitive ID accuracy. Relative to MViTv2-B, LATER improves OOD accuracy by 1.61 points, improves \(H\) by 1.54 points, and reduces \(\PD\) from 0.49 to 0.47.

The robustness scatter visualizes this trade-off in Figure~\ref{fig:id_ood_scatter}. It separates methods that fit the low-depression source distribution from methods that transfer better to high-depression OOD views. LATER moves the operating point upward relative to the strongest complete baseline, indicating better OOD recognition without a large ID sacrifice.

\begin{wrapfigure}{r}{0.45\linewidth}
  \vspace{-6pt}
  \centering
  \includegraphics[width=\linewidth]{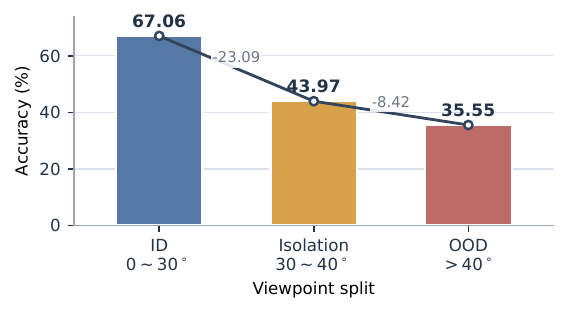}
  \caption{Viewpoint severity check. LATER accuracy decreases monotonically across the three splits.}
  \label{fig:isolation_diagnostic}
  \vspace{-6pt}
\end{wrapfigure}

The isolation-band diagnostic checks whether the removed samples behave as an intermediate viewpoint regime. In Figure~\ref{fig:isolation_diagnostic}, LATER accuracy drops from the low-depression ID split to the \(30\sim40^\circ\) isolation band, then drops further on the high-depression OOD split. This ordering supports the split design: the isolation band removes a genuine intermediate viewpoint regime rather than an arbitrary set of videos.

This diagnostic is not used as a main benchmark score. Instead, it checks whether the removed band behaves as a middle regime between ID and OOD. The intermediate accuracy supports this interpretation and reduces the chance that the final OOD gap comes from an abrupt sample-count artifact.

\subsection{LATER analysis and ablations}
\label{sec:later_ablation}

\begin{table}[htbp]
  \centering
  \begin{minipage}[t]{0.49\linewidth}
    \vspace{0pt}
    \caption{Effect of LoRA rank in LATER. Smaller ranks improve OOD robustness, while larger ranks recover ID accuracy.}
    \label{tab:rank_sweep}
    \centering
    \scriptsize
    \setlength{\tabcolsep}{2pt}
    \begin{tabular}{@{}lcccc@{}}
      \toprule
      Variant & \(\Acc_{ID}\) & \(\Acc_{OOD}\) & \(\PD\downarrow\) & \(H\uparrow\) \\
      \midrule
      \(r=2\) & 60.55 & 37.23 & \best{0.39} & 46.11 \\
      \(r=4\) & 62.42 & \best{37.97} & \best{0.39} & \best{47.22} \\
      \(r=8\) & 65.81 & 36.58 & 0.44 & 47.02 \\
      \(r=16\) & \best{67.06} & 35.55 & 0.47 & 46.47 \\
      \bottomrule
    \end{tabular}
  \end{minipage}
  \hfill
  \begin{minipage}[t]{0.49\linewidth}
    \vspace{0pt}
    \caption{Effect of re-centering strength \(\alpha\) in LATER. Larger \(\alpha\) improves OOD accuracy but lowers ID accuracy.}
    \label{tab:alpha_sweep}
    \centering
    \scriptsize
    \setlength{\tabcolsep}{2pt}
    \begin{tabular}{@{}lcccc@{}}
      \toprule
      Variant & \(\Acc_{ID}\) & \(\Acc_{OOD}\) & \(\PD\downarrow\) & \(H\uparrow\) \\
      \midrule
      \(\alpha=0.75\) & \best{67.84} & 35.19 & 0.48 & 46.34 \\
      \(\alpha=1.00\) & 67.06 & 35.55 & 0.47 & 46.47 \\
      \(\alpha=1.25\) & 66.35 & 35.61 & 0.46 & 46.35 \\
      \(\alpha=1.50\) & 65.71 & \best{36.13} & \best{0.45} & \best{46.62} \\
      \bottomrule
    \end{tabular}
  \end{minipage}
\end{table}

The rank sweep studies how LoRA adaptation capacity affects robustness, as shown in Table~\ref{tab:rank_sweep}. Increasing the rank steadily improves ID accuracy, from 60.55\% at \(r=2\) to 67.06\% at \(r=16\). OOD accuracy and \(H\), however, peak at smaller ranks. This supports the view that restricted source adaptation can reduce training-view shortcuts: a lower-rank update has less freedom to specialize all feature directions to the low-depression training distribution. We select \(r=16\) for the main table because it preserves the strongest ID accuracy while still improving the OOD/H trade-off over full fine-tuning baselines.

\begin{figure}[t!]
  \centering
  \includegraphics[width=0.96\linewidth]{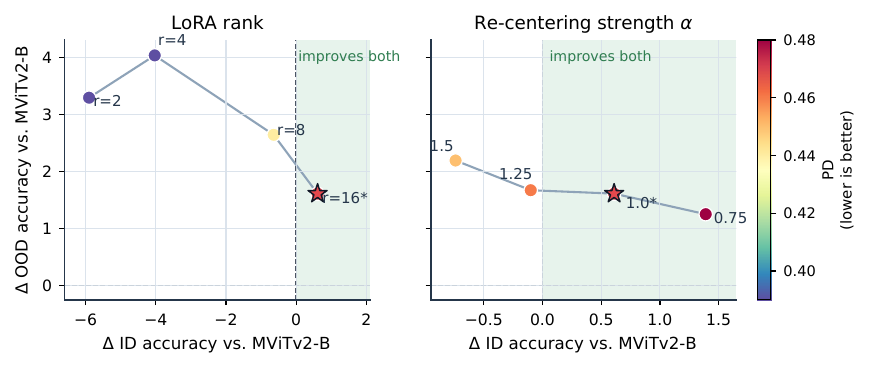}
  \caption{LATER trade-offs relative to MViTv2-B~\cite{Li2022MViTv2}. Points above and to the right of the dashed axes improve both ID and OOD accuracy. Color indicates \(\PD\), and stars mark the default \(r=16\) and \(\alpha=1.0\) settings.}
  \label{fig:later_sweeps}
\end{figure}

The strength sweep in Table~\ref{tab:alpha_sweep} shows a smooth ID/OOD trade-off. Larger \(\alpha\) applies a stronger correction and improves OOD accuracy, but it also moves ID features farther from the source-adapted classifier. We use \(\alpha=1.0\) as an ID-preserving default rather than the OOD-maximizing setting: it improves OOD accuracy over the strongest complete baseline while keeping ID accuracy competitive. Larger values such as \(\alpha=1.5\) further improve OOD accuracy and \(H\), but trade off more ID accuracy. Together, the two sweeps expose controllable robustness knobs in LATER: source adaptation capacity through rank and target correction strength through \(\alpha\), as summarized relative to the strongest complete baseline in Figure~\ref{fig:later_sweeps}.

\vspace{2mm}

\begin{table}[t!]
  \caption{LoRA-anchored projection versus global feature correction. Global Correction subtracts the full source-target centroid shift, while LATER removes only the component orthogonal to the LoRA subspace.}
  \label{tab:global_correction}
  \centering
  \footnotesize
  \setlength{\tabcolsep}{5pt}
  \begin{tabular}{@{}lcccc@{}}
    \toprule
    Variant & \(\Acc_{ID}\) & \(\Acc_{OOD}\) & \(\PD\downarrow\) & \(H\uparrow\) \\
    \midrule
    Global Correction & 63.29 & \textbf{36.23} & \textbf{0.43} & 46.08 \\
    LATER & \textbf{67.06} & 35.55 & 0.47 & \textbf{46.47} \\
    \bottomrule
  \end{tabular}
\end{table}
\vspace{2mm}

The projection ablation isolates the role of the LoRA anchor in Table~\ref{tab:global_correction}. Global Correction estimates a single source-target displacement and subtracts it uniformly across the full feature space. It achieves the highest OOD accuracy in this ablation, but it does so by substantially reducing ID accuracy. LATER sacrifices 0.68 OOD points relative to Global Correction, but recovers 3.77 ID points and achieves the best harmonic mean. This pattern supports the role of the LoRA anchor: projecting the displacement onto the orthogonal complement of the LoRA subspace preserves more of the source-adapted semantic directions while still correcting viewpoint-induced drift.

\section{Conclusion}
\label{sec:conclusion}

This paper studies out-of-viewpoint generalization in UAV action recognition. We introduced UAV-OVO, a benchmark that uses geometry-derived view scores to separate low-depression source views from held-out high-depression target views while keeping the action label space and matched ID/OOD test-set class distributions fixed. Experiments on UAV-OVO show that strong video recognizers can fit the training-view distribution yet lose substantial accuracy under viewpoint shift, indicating that aggregate benchmark accuracy can hide training-view shortcuts.

We also introduced LATER, a LoRA-Anchored Test-time Re-centering method for viewpoint-robust recognition. LATER first adapts the recognizer with a low-rank source update, then uses the learned LoRA subspace as a semantic anchor for gradient-free target correction. By re-centering only in the orthogonal complement of this subspace, LATER improves the ID/OOD balance without updating model parameters at test time. More broadly, UAV-OVO and LATER suggest that viewpoint robustness should be evaluated as an explicit deployment axis, and that parameter-efficient adaptation subspaces can provide useful structure for test-time correction.

{
\small
\bibliographystyle{unsrt}
\bibliography{ref}



}






\end{document}